\documentclass[letterpaper, 10 pt, conference]{ieeeconf}  

\usepackage{graphicx}
\usepackage{color}
\usepackage{amsmath}
\newcommand{\ie}{i.e.,\ }
\newcommand{\eg}{e.g.,\ }
\newcommand{\Reffig}[1]{Fig.~\ref{#1}}
\newcommand{\Refsec}[1]{Sec.~\ref{#1}}

\IEEEoverridecommandlockouts                              

\usepackage{cite}
\title{\LARGE \bf
TiEV: The Tongji Intelligent Electric Vehicle in the Intelligent Vehicle Future Challenge of China
}

\author{
Junqiao Zhao$^{*, 1, 2}$,
Chen Ye$^{1, 2}$,
Yan Wu$^{1, 2}$, 
Linting Guan$^{1, 2}$,
Lewen Cai$^{1, 2}$, 
Lu Sun$^{1, 2}$, 
Tao Yang$^{1, 2}$,\\ 
Xudong He$^{1, 2}$, 
Jun Li$^{1, 2}$, 
Yongchao Ding$^{1, 2}$, 
Xinglian Zhang$^{3}$, 
Xinchen Wang$^{1, 2}$, 
Jinlin Huang$^{1, 2}$,\\
Enwei Zhang$^{1, 2}$, 
Yewei Huang$^{3}$, 
Wei Jiang$^{1, 2}$,
Shaoming Zhang$^{3}$, 
Lu Xiong$^{4}$ and
Tiantian Feng$^{3}$
\thanks{This work is supported by the National Natural Science Foundation of China (No. U1764261), the Natural Science Foundation of Shanghai (No.kz170020173571) and the Fundamental Research Funds for the Central Universities (No. 22120170232)}
\thanks{$^{1}$The Key Laboratory of Embedded System and Service Computing, Ministry of Education, Tongji University, Shanghai
        {\tt\small zhaojunqiao@tongji.edu.cn}}%
\thanks{$^{2}$Department of Computer Science and Technology, School of electrics and Information Engineering, Tongji University, Shanghai}%
\thanks{$^{3}$School of Surveying and Geo-Informatics, Tongji University, Shanghai}%
\thanks{$^{4}$School of Automotive Studies, Tongji University, Shanghai}%
}

\begin{document}

\maketitle
\thispagestyle{empty}
\pagestyle{empty}

\begin{abstract}

TiEV is an autonomous driving platform implemented by the Tongji University of China.
The vehicle is drive-by-wire and is fully powered by electricity.
We devised the software system of TiEV from scratch, which is capable of driving the vehicle autonomously in urban paths as well as on fast express roads.
We describe our whole system, especially novel modules of probabilistic perception fusion, large-scale mapping and updating, the 1\textsuperscript{st} and the 2\textsuperscript{nd} planning and the overall safety concern.
TiEV finished 2016, and 2017 Intelligent Vehicle Future Challenge of China held at Changshu.
We show our experiences on the development of autonomous vehicles and discuss the future trends.

\end{abstract}

\section{INTRODUCTION}

The autonomous driving has long been seen as one of the ultimate solutions to transportation problems like the traffic jam and traffic accidents\cite{MontemerloBecker-2659,MaurerGerdes-2679}, and it is believed to be able to reform the way of traveling in our society\cite{Waymo-2681,Dmv-2589,BonnefonShariff-2667,MaurerGerdes-2679}.
In the past decade, the well-known DARPA grand challenge had proved its feasibility and demonstrated the technical frameworks for autonomous driving\cite{MontemerloBecker-2659,UrmsonAnhalt-2413,BergerRumpe-2611}.
Later led by universities as well as their industrial counterparts, autonomous driving researchers have witnessed a dramatic growth\cite{AeberhardRauch-2610,ZieglerBender-2227,CosgunMa-2678,BroggiDebattisti-2676,ShimChoi-2674,BuechelFrtunikj-2662}.
The latest prototypes, such as Waymo, have already shown their capability of driving more safely than human beings\cite{Waymo-2681,Dmv-2589}.
Nevertheless, in general, the autonomous driving technique is still in its early stages, especially when facing complex urban scenes where human drivers could easily interpret the traffic and act accordingly based on him/her experiences.

Sponsored by NSFC, China's similar event of DARPA urban challenge, the Intelligent Vehicle Future Challenge (IVFC) began from 2009\cite{IVFC-2016}.
In the last eight years, more than thirty universities, as well as companies, participated in this annual challenge, which is now recognized as the most influential event of the research and development of the autonomous driving in China.

\begin{figure}[h]
\centering
\includegraphics[height = 2.3in]{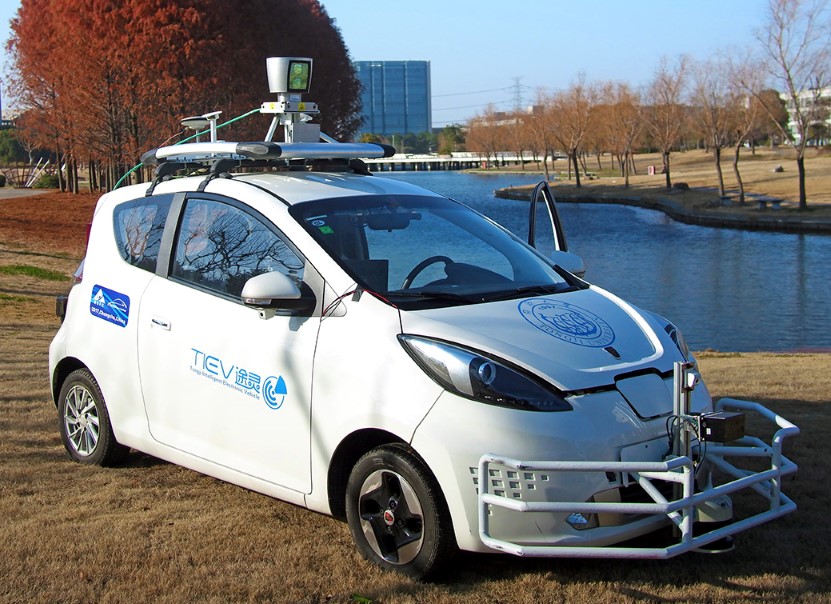}
\caption{
The TiEV autonomous driving platform
}\label{fig:car}
\end{figure}

As a newcomer of IVFC, Tongji Intelligent Electric Vehicle (TiEV) project\footnote{cs1.tongji.edu.cn/tiev} funded by the Tongji University was started in 2015.
A driverless prototype TiEV is built based on a modified electric vehicle (\Reffig{fig:car}).
It is equipped with vision sensors as well as laser scanners and an integrated localization system.
The computer systems are fused of two x86 IPCs and one embedded system.

%

Most of the software of TiEV are devised from scratch in C++ based on flexible cross-platform ZeroCM/LCM middleware\footnote{https://github.com/ZeroCM/zcm} and does not rely on off-the-shelf implementations, such as ROS and Autoware\footnote{https://github.com/CPFL/Autoware}.
Moreover, the TiEV software will also be made open-source in the future. 
TiEV proposed novel modules of probabilistic perception fusion (\Refsec{sec:perception}), large-scale mapping and updating (\Refsec{sec:mapping}), the 1\textsuperscript{st} and the 2\textsuperscript{nd} planning (\Refsec{sec:planning}), which will be detailed in the following sections.
The overall safety is of great importance in our systematic design (\Refsec{sec:safety}), which guarantees collision-free even if the planning module made the wrong decision.
TiEV participated in 2016 and 2017 IVFC and successfully managed to pass most of the tasks including simulated traffic, tunnels and blockages without human intervene.

The main contributions of this paper are listed as follow:
\begin{itemize}
\item We introduced the architectural design and innovational algorithms of our autonomous driving prototype TiEV.
\item We shared the experiences and our views on the competitions of IVFC of China and future development of autonomous driving vehicles.
\end{itemize}

\section{TIEV THE ARCHITECTURE}

\subsection{The vehicle}

TiEV is modified from Rowe E50 of SAIC motors (\Reffig{fig:car}).
The EPS and the motor can be controlled-by-wire through the CAN.
We install an electrohydraulic brake system (EHB) developed by the Institute of Intelligent Vehicles of Tongji University to enable the control-by-wire of the braking system.

\begin{figure}[b]
\centering
\includegraphics[height = 2.2in]{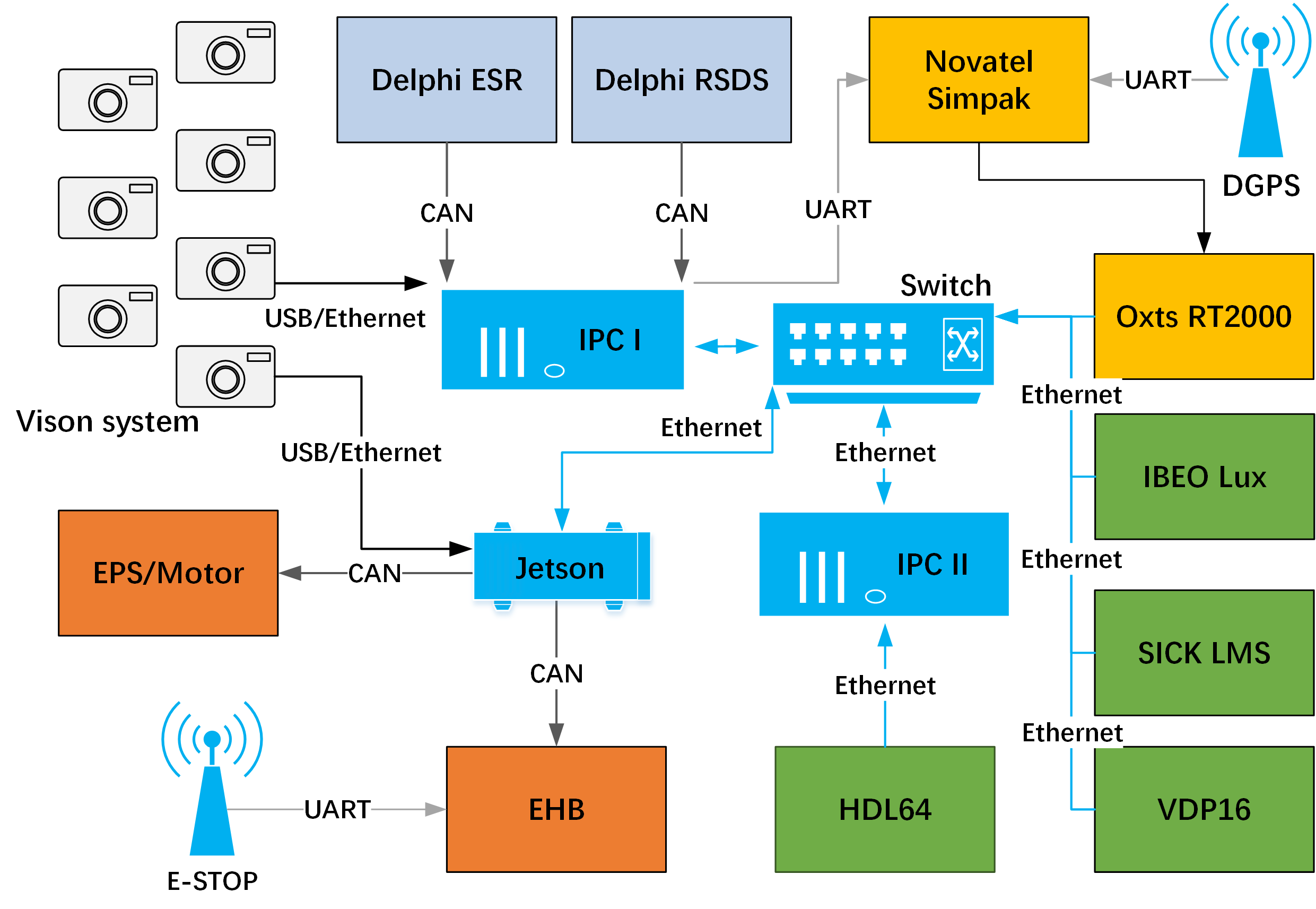}
\caption{
The system architecture of TiEV
}\label{fig:architecture_hw}
\end{figure}

\begin{figure}[t]
\centering
\includegraphics[height = 1.7in]{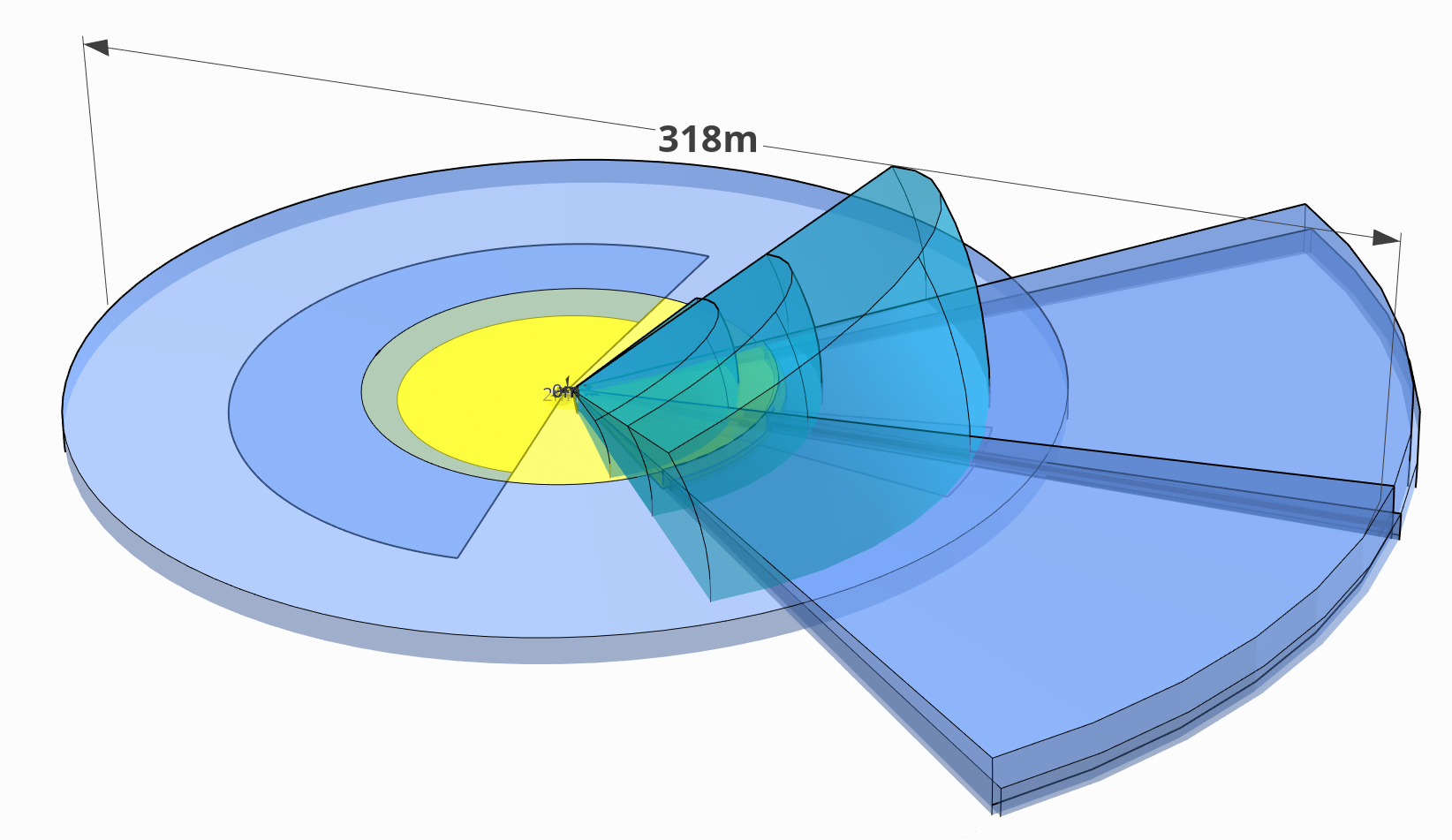}
\caption{
The perception coverage of TiEV
}\label{fig:perception_coverage}
\end{figure}

\Reffig{fig:architecture_hw} presents the overall architecture of the hardware system of TiEV.
We equipped seven vision sensors.
Two of the three forward-looking cameras compose a stereo vision system, and the other is connected to the embedded system for the detection task.
The four fish-eye cameras are calibrated to provide a top-down panorama view of about 10 meters by 10 meters (\Reffig{fig:vision_surround}).
The Velodyne HDL64 scanner is responsible for the segmentation of drivable area and the detection and tracking of moving obstacles such as pedestrians and vehicles.
An IBEO Lux4 and a Sick LMS511 scanner are installed to complement the blind-area of HDL64.
\Reffig{fig:perception_coverage} illustrates the spatial coverage of TiEV's sensors.

All the above sensors are calibrated interactively and transformed to the defined vehicle coordinate frame centered at the front axis, which is discovered to be helpful for obstacle avoidance.
A high precision DGPS+IMU system integrated by Novatel simpak6 GPS receiver and Oxts RT2000 IMU provides accurate localization information of about 10 centimeters in the outdoor environment.
We devise an EKF-based fusion method to integrate the vehicle kinematics and the IMU to keep the vehicle on track in GPS denied area such as in a tunnel.

Two of the three computers installed on TiEV are Advantech IPCs, and the other is an embedded system based on NVIDIA Jetson TX2.
TX2 is built with a Pascal GPU and the CAN communication capability.
As a result, a camera and the CAN bus are linked to TX2, on which the deep learning module and CAN actuation module are implemented.

\subsection{Modules overview}
\label{sec:module}
The software modules of TiEV are highly distributed, and communications between modules are decentralized.
Many similar systems have adopted this flexible and robust structure. 
We employed the ZeroCM middleware, which is lightweight, multifunctional and supports cross-platform.
The exact synchronization between modules is not required in our system.
A spatiotemporal stamp is introduced for the fusion of asynchronous information.
As a result, each module processes in its operation cycle, which is constrained by the upstream and downstream modules.

\begin{figure}[b]

\centering
\includegraphics[height = 2.2in]{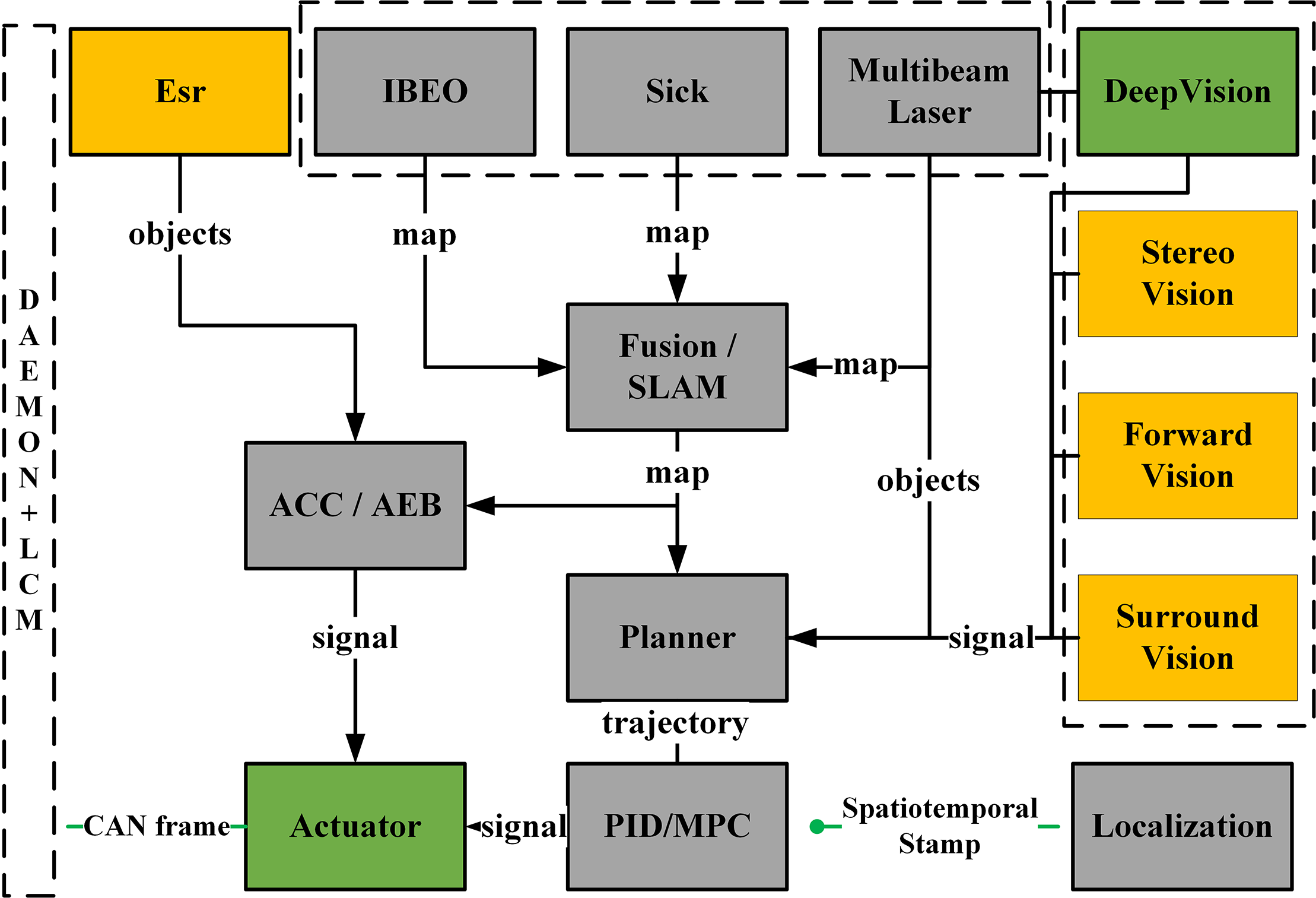}
\caption{
The software architecture of TiEV
}\label{fig:architecture_sw}
\end{figure}

\Reffig{fig:architecture_sw} presents the software modules and communications between modules. 
Different colors indicate the different computers on which modules are implemented, \ie{orange indicates \emph{IPC I}, gray indicates \emph{IPC II},  and green indicates \emph{Jetson}}.
This configuration is rooted from the considerations of the bandwidths and interface types of different sensors.
Messages are categorized into four classes, which are grid-based \emph{map}, tracked \emph{objects}, detected \emph{signal} and generated \emph{trajectory}.
The message of the spatiotemporal stamp is received by all modules.
The following sections will describe the featured modules in our system.

\section{PERCEPTION UNDER UNCERTAINTY}
\label{sec:perception}

Perception is the basis for autonomous systems.
Multiple sensor readings should be processed and fused into a unified representation for decision-making.
TiEV adopted a 2D grid-based representation for obstacles located within the decision region (80 meters by 30 meters and with the grid resolution of 0.2 meters).
However, sensor readings can be noisy thus we should fuse them in a probabilistic form. 
In this section, we will first address the modules for laser scanners and vision sensors.
Then the fusion methods will be explained.

\subsection{Laser perception}
\subsubsection{Multibeam Laser}
This module processes the 3D point clouds send from HDL64. 
We implement obstacle segmentation, classification, and tracking consecutively.

The segmentation is conducted in two steps:
Firstly, the average of $n$ lowest $z$ values of points in an upsampled grid cell\footnote{We adopt 4 or 5 times upsampling, which is about 1 meter resolution.} is counted as $z_{min}$.
All the points in the upsampled grid cell that are higher than $z_{min}$ by a certain amount are classified as obstacle points.
This step results in a coarse segmentation of obstacles, which can better preserve flat surfaces such as roofs of cars than directly using the finer grid.
In the next step, we traverse each of the original grid cells.
The cell which contains obstacle points located within the vertical span of the vehicle is marked as an obstacle.
Therefore influence from tree branches can be eliminated.

%

\begin{figure}
\centering
\includegraphics[height = 2.3in]{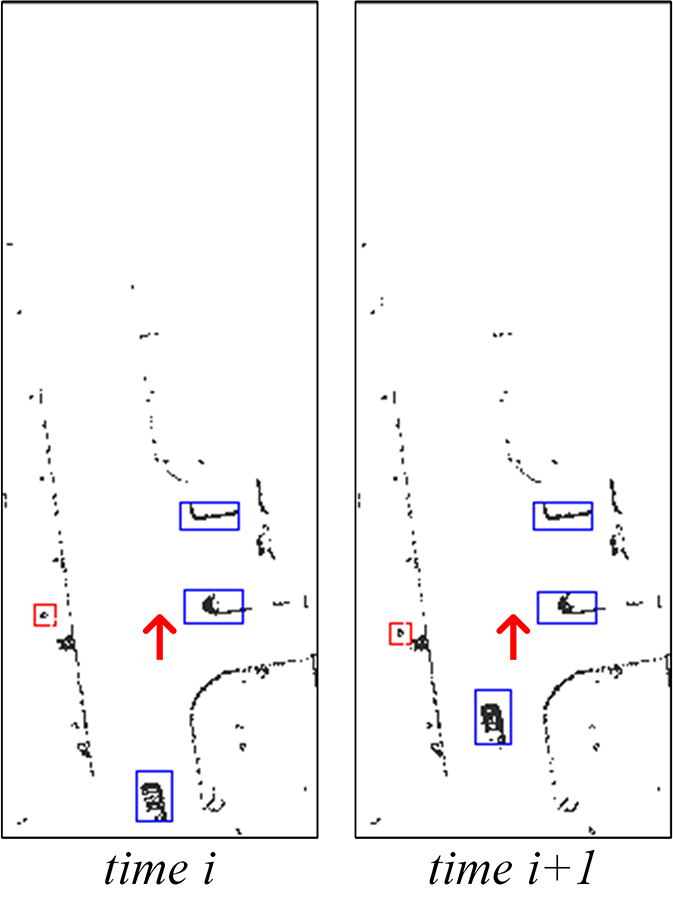}
\caption{
The segmentation, classification and tracking of static and dynamic obstacles based on HDL64 (blue boxes indicate vehicles, red boxes indicate pedestrians, the red arrow represents the ego pos)
}\label{fig:hdl64}
\end{figure}

In the meanwhile, we classify each of the obstacle clusters based on a multiboost classifier proposed in \cite{Semisupervise}, which results in cars, bicycles, and pedestrians.
Kalman filters are then created for each object for filtering and predicting their movements in adjacent frames (\Reffig{fig:hdl64}).
The detection and tracking processes are further constrained by the \emph{Forward vision} module. 

\subsubsection{Sick and IBEO}
These sensors are mainly used as complements to HDL64, especially in the near and far front of the vehicle.
We compensate the extrinsic parameters of both sensors based on $pitch$ readings from IMU and project the points onto the grid map.
Afterward, all the grid maps generated from laser perception modules will be fused in the \emph{Fusion} module (\Refsec{sec:fusion}).


\subsection{Visual perception}
\subsubsection{Forward vision} 
\begin{figure}
\centering
\includegraphics[height = 2.3in]{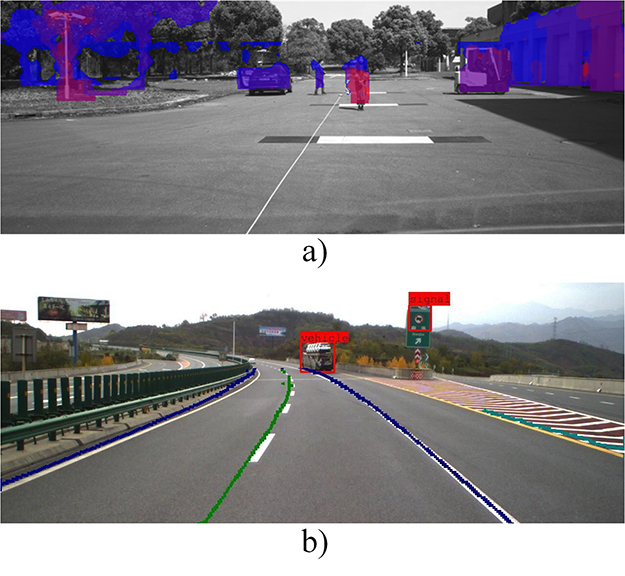}
\caption{
The results of \emph{forward vision module} (a) shows the depth map from stereo vision; b) shows the signal and vehicle detection and lane segmentation from deep vision)
}\label{fig:vision_fw}
\end{figure}

We obtain rich visual information from three Basler Ace cameras mounted behind the windshield and above the rear-view mirror of the vehicle. 
Two of them compose a stereo rig for coarse depth vision (\Reffig{fig:vision_fw} a)) and the other is used for detection.
We trained a model based on YOLO2 to detect cars, pedestrians, and traffic signs \cite{2016arXiv161208242R}.
The results are shown in \Reffig{fig:vision_fw} b).
2D boxes of cars and pedestrians are further mapped to 3D point cloud in \emph{Multibeam Laser} module by inverse perspective projection based on the camera-laser calibration\cite{unnikrishnan2005fast}.
They can provide semantic references for object detection and tracking in the \emph{Multibeam Laser} module. 

\subsubsection{Surround vision}
\begin{figure}
\centering
\includegraphics[height = 1.5in]{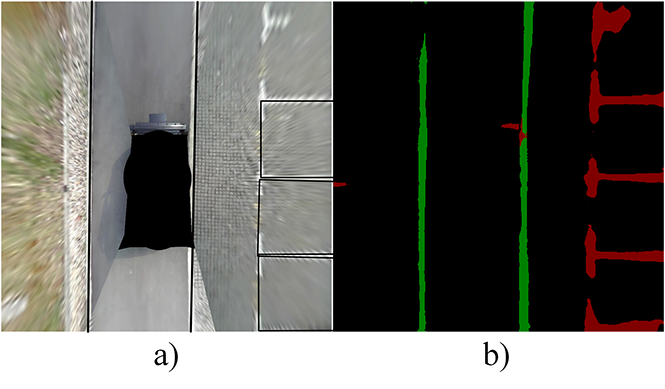}
\caption{
The road marker extraction based on the surround vision system (a) shows the detected markers overlapped to the original panorama image; b) shows the counterpart segmented results)
}\label{fig:vision_surround}
\end{figure}

Based on the panoramic view (\Reffig{fig:vision_surround} a)) fused by four fisheye cameras, TiEV introduces a VH-stage to HFCN to robustly segment parking slots and lane markings under various scene and illumination conditions \cite{Yang2018Semantic}.
The results are shown in \Reffig{fig:vision_surround} b).

\subsection{Synchronization and Fusion}
\label{sec:fusion}
Notwithstanding our modules do not require hard synchronization as mentioned in \Refsec{sec:module},
multiple perceptual modules observe the surroundings in different frequencies.
Their observation results should be aligned and fused.

%

Pure timestamps based systems require a universal extrapolator to interpolate poses.
Alternatively, we explicitly attached a spatiotemporal stamp $(t, x, y, \theta)$ published by the \emph{localization} module to messages in each sensory module once a measurement is made.
The pose information will then be directly used in the fusion stage for transforming the observations to the current pose.
The overall mismatch is about several centimeters, which is due to the millisecond-level time delay in receiving the stamped message and is satisfied with driving even at relatively high speed.
The stamp also provides a temporal constraint for the fusion, which ensures the latest observation is adopted.

Lacking context information as multibeam laser scanners offered, false alarms and noises generated during severe maneuvers of the vehicle, such as an emerge brake, cannot be easily removed for "sparse-beam" laser scanners, \eg{ 1, 4 or 8 beams}.
Therefore, a probabilistic fusion method of multiple sensors is employed in our system.
The fusion is performed in the overlapping region of different laser measurements defined by the intersections of their field of views in $xy$ plane.
A 2D virtual scan is firstly generated from \emph{Multi-beam Laser} module to be aligned with \emph{IBEO} and \emph{Sick}.
The occupancy-based representation is then adopted for the fusion of obstacles, in which the Odds of a cell is derived by:
\[
\frac {p(x \mid z_{{1:t}})} {p(\bar x \mid z_{{1:t}})} = \prod_{k=1}^{n} \frac {p(x \mid z_{{t}}^{{k}})} {p(\bar x \mid z_{{t}}^{{k}})} \cdot \frac {p(x)} {p(\bar x)} \cdot \frac {p(x \mid z_{{1:t-1}})} {p(\bar x \mid z_{{1:t-1}})}
\]
where $p(x)$ is the state of being obstacle in one grid cell and  $p(\bar x)$ is its complementary.
$z^k$ indicate measurements of sensors $k=1,\dots,n$, represented by \emph{maps} from each sensory module.

\begin{figure}
\centering
\includegraphics[height = 1.9in]{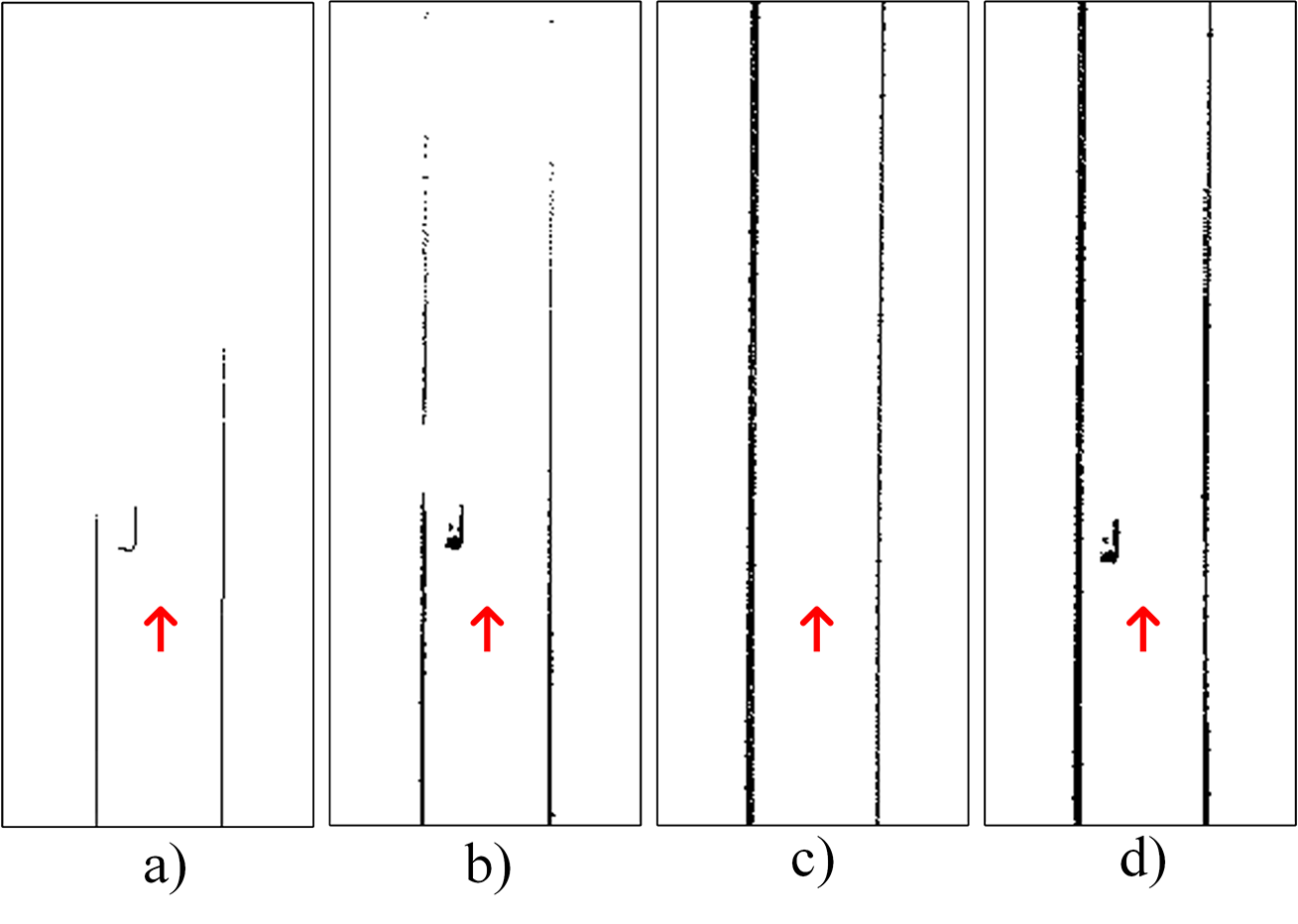}
\caption{
The probabilistic fusion of laser detections (a) shows the fused drivable area from multiple sensors; b) shows the segmented obstacle from \emph{Multi-beam laser}, which needs to be complemented by a) in narrow paths; c) shows the historical map of drivable area; d) shows the finally fused map for planning)
}\label{fig:fusion}
\end{figure}

We choose a belief threshold of 0.75 to conservatively extract the maximum likelihood map (\Reffig{fig:fusion} a)).
This fused map represents the drivable area closed to the vehicle, covering the blind-area of HDL64.
Influences of noises, as well as false alarms, generated from SICK or IBEO Lux sensors,  are drastically abbreviated.

Finally, $map$ of segmented obstacle from \emph{Multi-beam Laser} module (\Reffig{fig:fusion} b)) and the historical $map$ from \emph{Mapping} modules (\Reffig{fig:fusion} c)) (described in \Refsec{sec:mapping}) are merged to the grid cells of state $unknown$ ($p(x) = 0.5$) generated from the previous step, which breads the final fused map used in the \emph{planner} module (\Reffig{fig:fusion} d)).

\section{LARGE-SCALE MAPPING AND UPDATING}
\label{sec:mapping}

Planning demands the historical map of the static driving environment \eg{when the view of sensors are occluded}.
However, the existing methods for generating HD driving map are usually labor intensive and expensive.
Based on the local probabilistic fusion, we extend it to mapping the historical map of the whole driving environment fully automatically. 

%

\begin{figure}
\centering
\includegraphics[height = 1.5in]{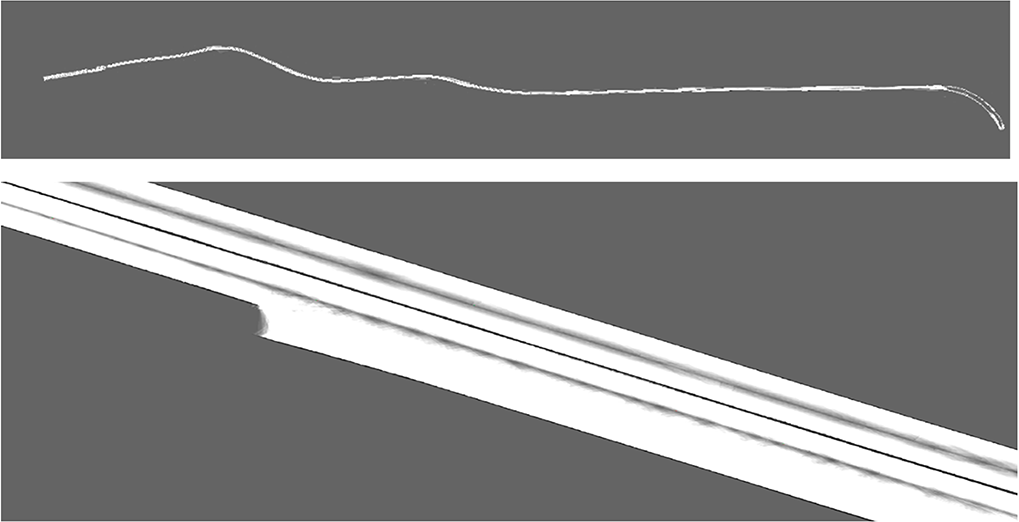}
\caption{
The map of drivable area of 12km's long
}\label{fig:slam}
\end{figure}

The state-of-the-art SLAM system, \eg{Cartographer} \cite{HessKohler-2479}, stored all the local maps and scans in memory, which can be burdensome when used in large-scale mapping.
Alternately, we take the usage of R-tree to index all the locally fused maps.
Only the visible portion of the map will be kept in the memory while others are streamed out.
In this way, we could map almost an area of arbitrary size.
\Reffig{fig:slam} shown a 12 km long path mapped using our method in 30 minutes.
The memory print is bounded continuously by around 150 MB.
Moreover, the time efficiency of maintaining and searching are guaranteed by R-tree.

\begin{figure}
\centering
\includegraphics[height = 1.7in]{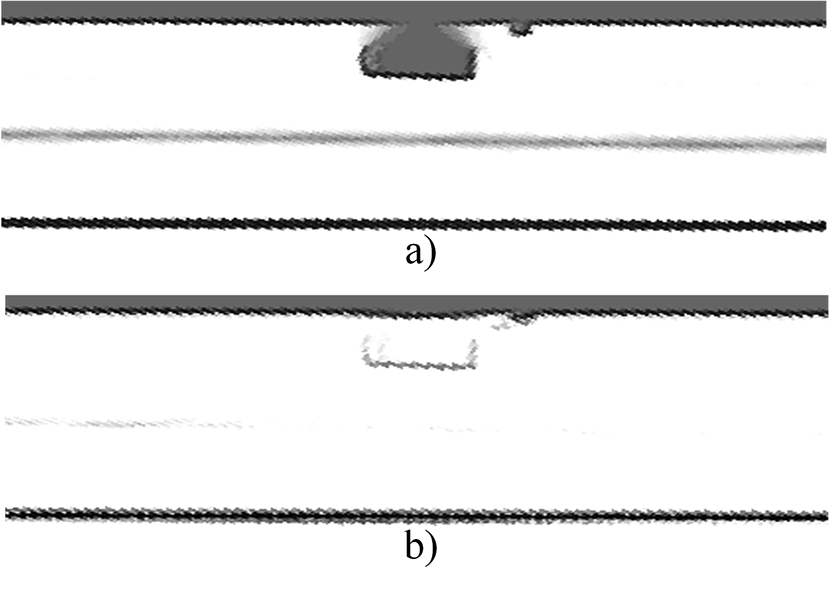}
\caption{
The updating of the map of drivable area (a) shows a previously build map with a car parking on the road side; b) shows the updated result of the area in another day, where the influence of the car has been eliminated)
}\label{fig:update}
\end{figure}

In practice, the mapping of a large area could not be realized in one go, because of the limits of the battery life or the traffic conditions during mapping.
Therefore, we implemented an incremental mapping strategy thanks to the streaming design.
When one mapping process finished, all the local maps are streamed onto disk.
In the next run, our mapping module could load the surrounding previously built maps and restore the mapping process.
This mechanism brings the extra benefit of automatically map updating.
We allow the overlapping between local maps, both spatially and temporally.
When retrieving the visible maps during driving, we fused the surrounding overlapping local maps based on a weighted averaging strategy.
The probability of a fused grid cell is derived from:
\[p(x) = \frac {1}{n} \Sigma_{i=1}^n w_i \cdot p(x_i)\]
where $w_i$ is the weighting for the $ith$ local map, which is given according to the date of acquisition.
\Reffig{fig:update} shows a local map with a car parking on the roadside a), which is updated by the second round mapping in another day b).

\section{THE 1\textsuperscript{ST} AND 2\textsuperscript{ND} PLANNING}
\label{sec:planning}
We termed the path planning as the 1\textsuperscript{st} planning and the trajectory planning as the 2\textsuperscript{nd} planning because of their temporal relation.
The 1\textsuperscript{st} planning is only triggered when the current path cannot be continued, such as the road is blocked.
The 2\textsuperscript{nd} planning operates in real-time when the vehicle is moving according to the path.
We implement the less-frequent 1\textsuperscript{st} planning in an open source spatial database system and proposed a novel unified 2\textsuperscript{nd} planning for both the structured and unstructured environment. 

\subsection{Path planning - The 1\textsuperscript{st} planning}
The 1\textsuperscript{st} planning is based on HD maps captured by using our vehicle and edited using QGIS\footnote{https://www.qgis.org/}.
All the lanes and intersections are sampled and topologically connected to form a lane-based road network.
The road network also records path-related information such as the speed limits, the right of lane-change.
We choose to manage the HD map with a spatial database rather than map files because such a map is mostly constant and infrequently updated.
And the most important, such an HD map should be able to be accessed by multiple users simultaneous.

\begin{figure}
\centering
\includegraphics[height = 1.6in]{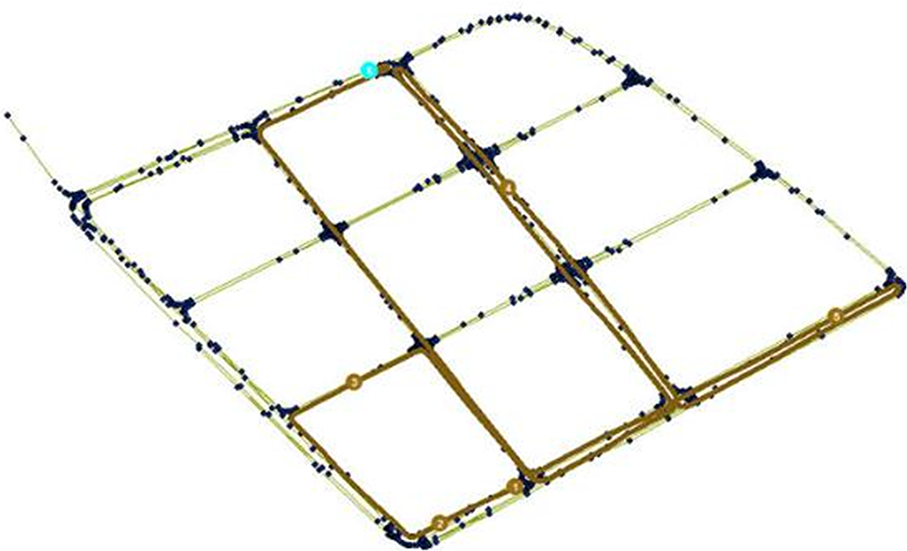}
\caption{
The path planning results on the lane-based road network using pgRouting
}\label{fig:path_planning}
\end{figure}

We adopt the open-source spatial database PostGIS\footnote{http://www.postgis.net/}, and use pgRouting\footnote{http://pgrouting.org/} to perform the shortest path finding (\Reffig{fig:path_planning}).
This database-based implementation has a high performance and provides multiple accesses from multiple autonomous vehicles if the database is accessed remotely on a server. 

%


\subsection{Trajectory planning - The 2\textsuperscript{nd} planning}
\label{sec:2ndplanning}
\begin{figure}
\centering
\includegraphics[height = 2.4in]{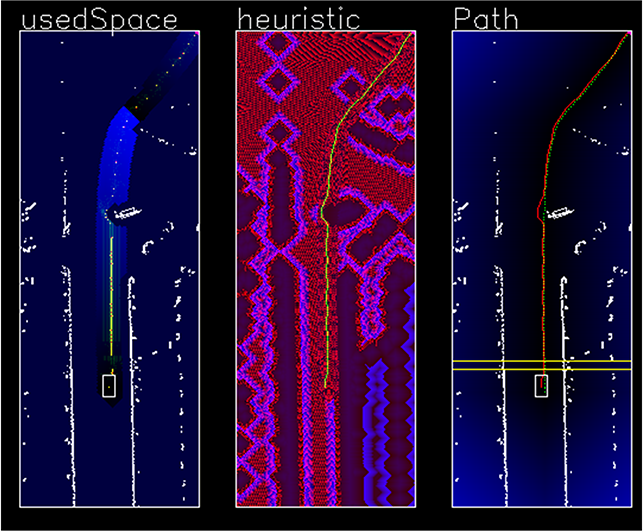}
\caption{
The interface of TiEV's \emph{planner} module, which shows a trajectory (the red dotted line) generated based on the obstacles as well as the path given by the 1\textsuperscript{st} planning (the green dotted line)
}\label{fig:planner}
\end{figure}

TiEV introduces a unified planning module for both the structured and unstructured driving environment.
An enhanced real-time A* algorithm is proposed to find the optimal path on the grid-based representation.
We model the lanes, the static and dynamic obstacles, the parking space and the path from the 1\textsuperscript{st} planning as weightings according to unified weighting police based on the breadth-first search (\Reffig{fig:planner} middle).
Optimizations, including simplifying the collision detection, discretization of angle states and pre-calculating the kinematics-aware heuristics, are proposed to bound the time expense of planning to within 20 to 80 milliseconds.

As a result, our unified planner could greatly simplify the finite state machine in our decision-making engine.
Moreover, it offers more flexible and intelligent planning than conventional trajectory generation methods while TiEV is running on complicated urban roads.

\section{THE SAFETY CONCERN}
\label{sec:safety}
Safety design is the highest priority of TiEV system.
We regard the safety of all traffic participants on the road as well as the vehicle itself.
The implementations will be described in car behavior and system design respectively.

\subsection{Safety concern of TiEV's behavior}
One of the primary goals of autonomous driving is to minimize the rate of traffic accidents.
To our understanding, the top safety guarantee of an autonomous vehicle is to prevent active collision of the vehicle to any traffic participants, including pedestrians, cyclists, other vehicles, and infrastructures.
This fatal behavior is tightly constrained in TiEV by the introduction of a dual ACC/AEB implementation.

The ACC/AEB function is implemented within two modules.
The \emph{2\textsuperscript{nd} planning} module calculates a safe speed $v_{safe}$ in real time based on the distance to obstacles on the referenced path according to the following equation  \cite{Chenyi-2372}:
\[
v_{safe}(t)=v_{max}(1-exp(-\frac{c}{v_{max}}dist(t)-d))
\] 
where $v_{max}$ is the highest permitted speed, $dist(t)$ is the distance between ACC/AEB target and current vehicle, $c$ and $d$ are model parameters.

\begin{figure}
\centering
\includegraphics[height = 2.3in]{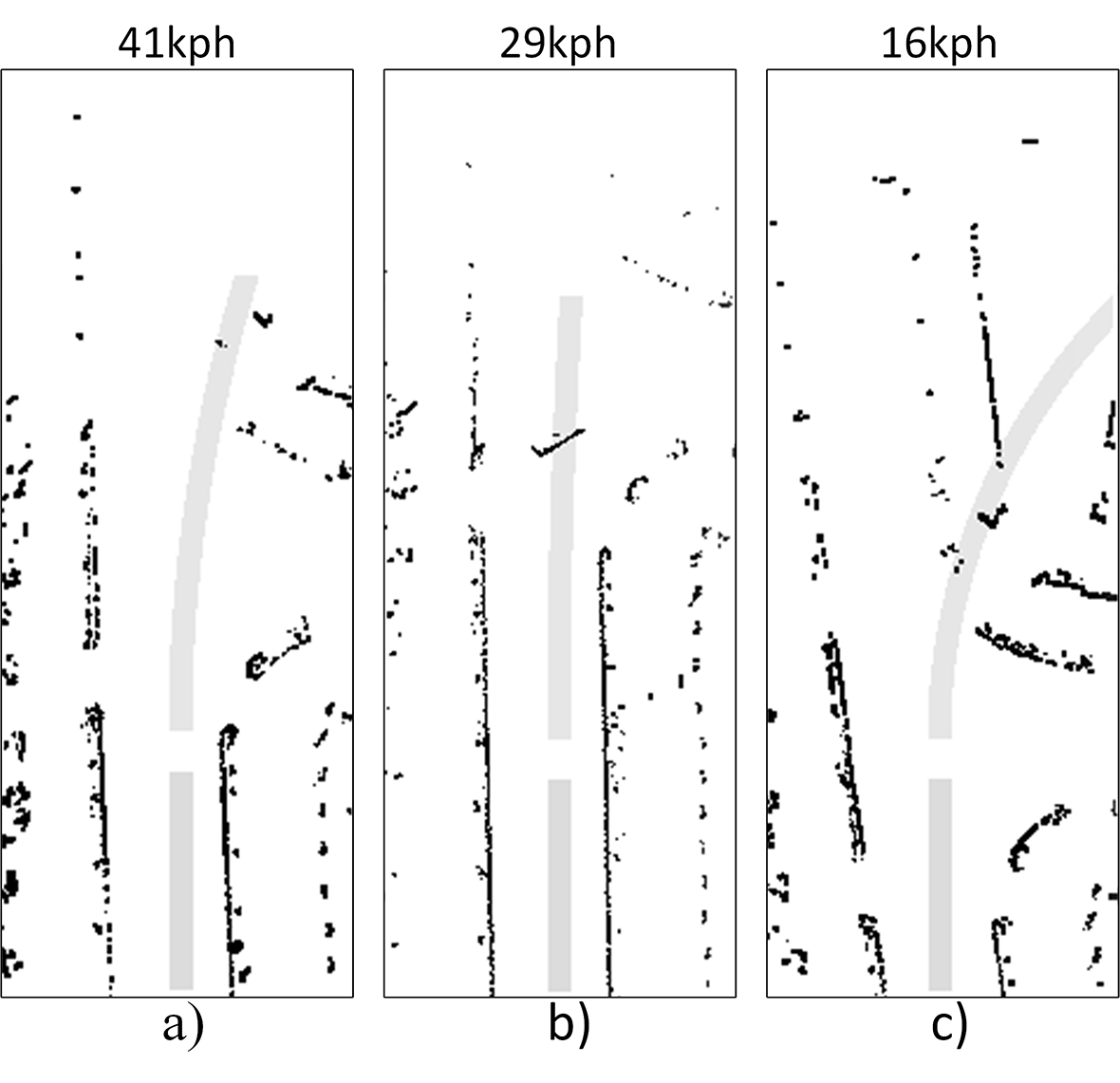}
\caption{
The safe speed calculation in \emph{ACC/AEB} module, where the gray zone represents the predicted path based on current control states
}\label{fig:acc}
\end{figure}

In the meantime, an independent \emph{ACC/AEB} module is introduced and acts as a shadow planner, which bypasses the main planning module and directly communicates to the actuator module (\Reffig{fig:architecture_sw}).
This module receives both the observed obstacle maps from each of the sensory modules, as well as the historical maps sent by the perception fusion module.
The runtime steering angle and velocity are integrated to estimate a future trajectory according to current control states (\Reffig{fig:acc}).
A safe speed is then calculated according to the previous equation.

The target speed for actuator will then be restricted by both the speed limits extracted from the map and the above two safe speeds.
We run the ACC/AEB module on two different computers to increase the redundancy.
So we seldom collide with any static or moving objects during our experiments.
\subsection{Safety concern of TiEV's system}
As an autonomous car will eventually carry families, driving on real roads every day\cite{MaurerGerdes-2679}, any system faults cannot be tolerated, especially those related to the core functionality.
Although TiEV is designed as an experimental prototype, we make the design to fulfill the systematic safety requirements.

The daemon modules that listen to heartbeats of all other modules are implemented redundantly.
On a local computer, a daemon module tries to restart local modules that were are no longer sending out heartbeats or sending out heartbeats with frozen spatiotemporal stamps.
A forked child thread of itself can also monitor the daemon module.
Remotely, the heart beats of daemon modules are also monitored by each other from separate computers.
Once a daemon module is judged as a failure, it means either the computer or the networking service of that computer has failed.
In this case, TiEV will try to stop the vehicle immediately.
To guarantee the robust communication between computers and the vehicle, we also introduce redundant CAN communication interfaces.
The default CAN control messages are sent by modules on the TX2 embedded system.
However, another CAN interface is installed on one of the IPCs as the backup.
Both computers could monitor and send CAN messages independently.

In practice, TiEV system has an extremely low probability of failure. 
One can even introduce redundancy to computers to further decrease the risks.

\section{EXPERIENCES GAINED FROM THE FUTURE CHALLENGE}



\begin{figure}
\centering
\includegraphics[height = 1.2in]{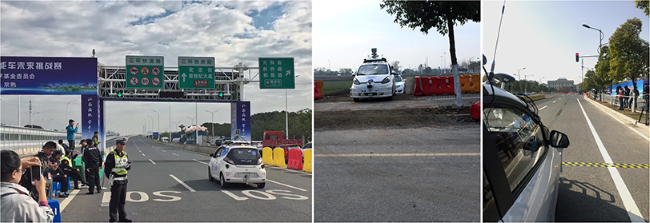}
\caption{
TiEV in the IVFC
}\label{fig:challenge}
\end{figure}

The on-road competition of the IVFC is composed of two events, \ie{the express road competition (around 12km with more than 10 tasks) and the urban road competition (around 3km with more than 20 tasks)} (\Reffig{fig:challenge}).
The exact task points of both events are not released until 30 minutes before the competition.
As a result, the participants have to build their roadmap in advance.
In the competitions, vehicles have to recognize various situations on the road, \eg{signals, blockages, tunnels, pedestrians, other vehicles} and behave appropriately.
The final score will be given based on evaluations of the task achievements, the traffic violations, and the time costs.
TiEV ranked the 11th (out of 25) and the 13th (out of 28) in the IVFC 2016 and 2017 respectively. 

Our main lessons gained can be laid in three-folds.
The first is the lack of comprehensive perception ability.
TiEV could recognize traffic signals, three kinds of traffic participants as stated in \Refsec{sec:perception}, but it treats others only as obstacles.
This cause problem when coming across specific scenes containing barriers made by \eg{reflective triangles or cones}, which warn the driver of blockages located ahead.
A human driver would interpret the scene based on their meanings rather than regarding them as generic obstacles.
As a result, our detector should be enhanced by learning an enriched set of traffic signs and objects.

%

Secondly, the tight coupling of autonomous driving and high precise lane-based map can be problematic if the map is erroneous because of, \eg{simply out of date}.
The using of the lane-based map also demands highly precise localization.
Our planning method is designed not to follow the lane-based path precisely.
TiEV treats the path as one of the references as the detected lanes and obstacles, and decide the best planning goal for the 2\textsuperscript{nd} planning.
Nevertheless, a coarse path shifted meters from the correct position would still cause the problem, especially when the planning goal cannot be decided wisely.
In contrary, human drivers could drive according to inaccurate maps or even with only directional instructions.

%

Finally, the optimal searching nature of the 2\textsuperscript{nd} planning, which keeps on trying to find the best trajectory will abruptly turn the wheel when the current trajectory is deformed. 
This behavior results in an agile but uncomfortable and riskful riding experience.
To abbreviate this effect, we introduce a piecewise planning strategy that keeps a local window of trajectory constant and concatenates the dynamic trajectory smoothly at the point beyond the look-ahead region.
In practice, this method successively stabilizes the vehicle when driving up to 60kph (the highest speed limit of IVFC).



\section{CONCLUSION REMARKS}
We realized that the ability of TiEV is still far from what we demanded for practical applications on complicated urban roads.
The system could already cope with many scenarios and drive the vehicle safely.
Nevertheless, it still cannot match up to human drivers in respect of adaptivity to environmental variations and robustness when facing noises. 

%

At present, deep learning-based methods have already proved their abilities to detect and segmenting different types of objects from images almost in real time \cite{2016arXiv161208242R}. 
However, such perception cannot still understand the characteristics of objects or their relations, \eg{we can tell from a new driver in front of us based on his/her reactions to certain traffic conditions.}
Many of these in-depth understanding of traffic scenes contribute to appropriate driving behavior.
This can only be realized with the help of the modeling of driving experiences. 
SLAM community intensively studies the memorization of the static driving environment.
However, the modeling of the contextual semantics of driving experiences, such as the relations and interactions between objects in a driving environment, is still an open question. 

%

Besides, the online perception burden can easily surpass the onboard computing resources.
Human driver relieves this burden by emphasizing on specific objects according to current driving intentions, which is known as the attention mechanism. 
It calls for a tighter coupling between the planning and the perception functions in the future.

Moreover, human drivers usually do not have to behave "optimally" as algorithms do.
Planning optimization should, therefore, be relaxed for temporally suboptimal solutions and should be regularized to react robustly to disturbances.

At last, we argue that the technical route of autonomous driving is still under drastic evolution.
Differentiations of implementations worldwide will eventually benefit to the maturation of the autonomous driving systems.

\addtolength{\textheight}{-12cm}   





\section*{ACKNOWLEDGMENT}
Special thanks to Deyi Li, Wei Han, Changzhu Zhang, Lifeng An, Dan Hai, Jing Zhu and Peizhi Zhang for the provided help and discussions during the implementations.



\bibliographystyle{IEEEtrans}
\bibliography{IEEEabrv,tiev}

\end{document}